\documentclass[letterpaper]{article} 
\usepackage{aaai2026}  
\usepackage{times}  
\usepackage{helvet}  
\usepackage{courier}  
\usepackage[hyphens]{url}  
\usepackage{graphicx} 
\usepackage{xcolor}
\usepackage{natbib}  
\usepackage{caption} 
\usepackage{algorithm}
\usepackage{algorithmic}
\usepackage{booktabs}

\usepackage{newfloat}
\usepackage{listings}
\DeclareCaptionStyle{ruled}{labelfont=normalfont,labelsep=colon,strut=off} 
\floatstyle{ruled}
\newfloat{listing}{tb}{lst}{}
\floatname{listing}{Listing}
\title{Symphony-MoE: Harmonizing Disparate Pre-trained Models into a Coherent Mixture-of-Experts}
\author{
Qi Wang\textsuperscript{$\ddagger$,$\dagger$,$\natural$,$\flat$}, Hanyang Peng\textsuperscript{$\dagger$}, Yue Yu\textsuperscript{$\dagger$}\thanks{Corresponding Author}}

\affiliations{ \textsuperscript{$\ddagger$} Key Lab of Intelligent Information Processing, Institute of Computing Technology, Chinese Academy of Sciences
 \\
 \textsuperscript{$\dagger$}Peng Cheng Laboratory \ \ \ \  \textsuperscript{$\natural$}University of Chinese Academy of Sciences
 \\
 \textsuperscript{$\flat$}State Key Laboratory of AI Safety, Institute of Computing Technology, Chinese Academy of Sciences
 \\
 wangqi245@ucas.mails.ac.cn
}

\usepackage{bibentry}

\begin{document}

\maketitle

\begin{abstract}
Mixture-of-Experts (MoE) models enable scalable performance by activating large parameter sets sparsely, minimizing computational overhead. To mitigate the prohibitive cost of training MoEs from scratch, recent work employs \textit{upcycling}, reusing a single pre-trained dense model by replicating its feed-forward network (FFN) layers into experts. However, this limits expert diversity, as all experts originate from a single pre-trained dense model. This paper addresses this limitation by constructing powerful MoE models using experts sourced from multiple identically-architected but disparate pre-trained models (e.g., Qwen2.5-Coder and Qwen2). A key challenge lies in the fact that these source models occupy disparate, dissonant regions of the parameter space, making direct \textit{upcycling} prone to severe performance degradation. To overcome this, we propose Symphony-MoE, a novel two-stage framework designed to \textit{harmonize} these models into a single, coherent expert mixture. First, we establish this harmony in a training-free manner: we construct a shared backbone via a layer-aware fusion strategy and, crucially, alleviate parameter misalignment among experts using activation-based functional alignment. Subsequently, a stage of post-training coordinates the entire architecture. Experiments demonstrate that our method successfully integrates experts from heterogeneous sources, achieving an MoE model that significantly surpasses baselines in multi-domain tasks and out-of-distribution generalization.
\end{abstract}


\section{1 Introduction}
\begin{figure}[t]
\centering
\includegraphics[width=0.45\textwidth]{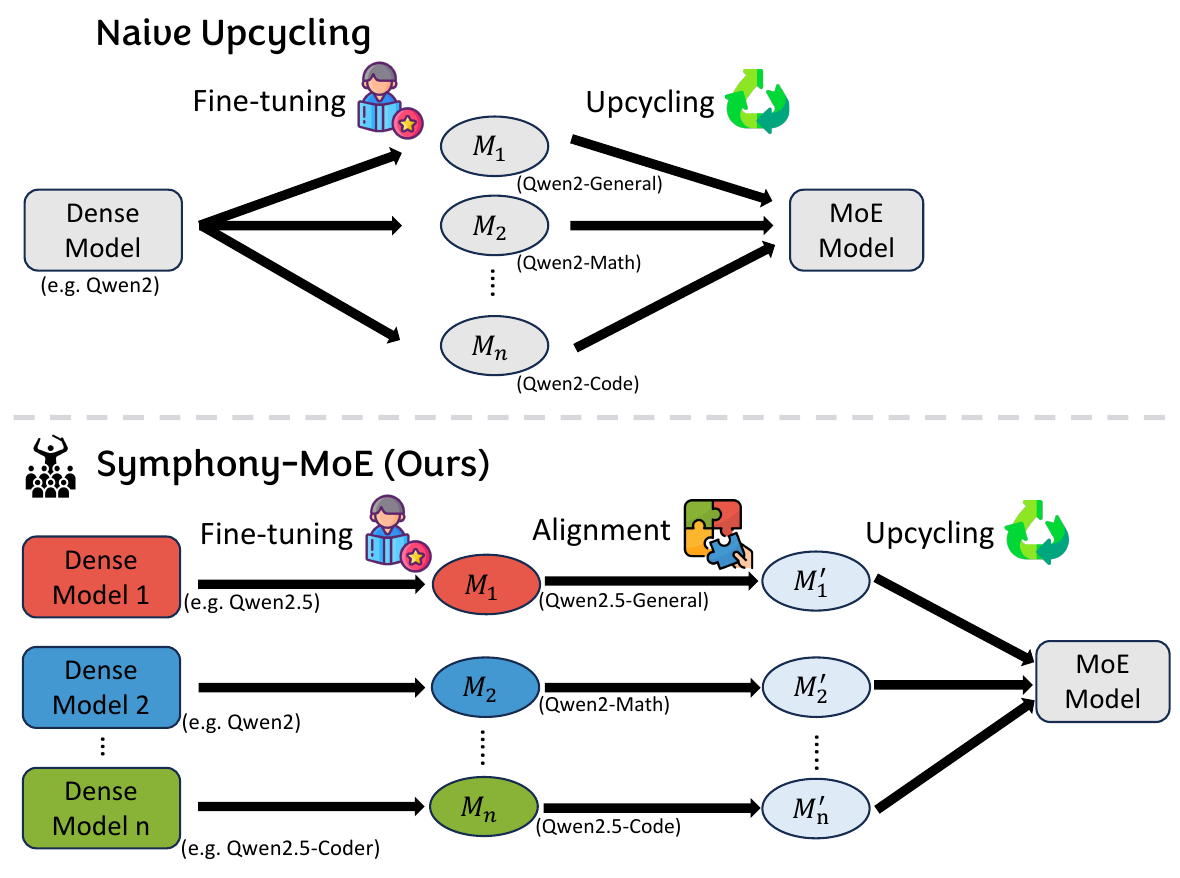} 
\caption{Comparison between the workflow of naive upcycling and ours. }
\label{fig_teaser}
\end{figure}

Large language models (LLMs) have demonstrated remarkable progress in recent years, with the Mixture-of-Experts (MoE) architecture emerging as a key paradigm for efficient scaling~\citep{shazeer2017outrageously}. By dynamically routing computations to sparse subsets of parameters—known as ``experts"—MoE models achieve vast total capacity while maintaining a near-constant floating-point operations (FLOPs) count per forward pass.

Despite their potential, training MoE models from scratch requires prohibitive computational resources and massive datasets. To mitigate this, academia and industry have explored \textit{upcycling} techniques~\citep{komatsuzaki2022sparse, zhu2024llama}. These methods typically expand a pre-trained dense model’s feed-forward network (FFN) layers into MoE layers, initializing each expert as a replica of the original FFN. Subsequent continual training then encourages expert specialization, with studies confirming that diverse data subsets can effectively differentiate replicated experts~\citep{sukhbaatar2024branch}.

However, existing upcycling approaches such as ``Branch-Train-Mix"~\citep{sukhbaatar2024branch} share a fundamental limitation: reliance on a single pre-trained checkpoint (see Figure~\ref{fig_teaser}). Consequently, all experts originate from an identical parameter space, regardless of subsequent specialization. This ``single-origin" constraint intrinsically caps expert diversity. In practice, a large number of high-quality, identically architected models—specialized through distinct pre-training objectives, data corpora, or versions—exist, such as the dialogue-optimized Llama2-Chat~\citep{touvron2023llama} and the code-centric Code Llama~\citep{roziere2023code}. These models capture unique knowledge and capabilities within their domains, occupying valuable yet heterogeneous regions of the parameter space. Directly leveraging such specialized models as distinct experts offers a promising avenue for constructing more powerful and versatile MoEs, but remains infeasible under current frameworks.

\begin{figure*}[t]
\centering
\includegraphics[width=0.98\textwidth]{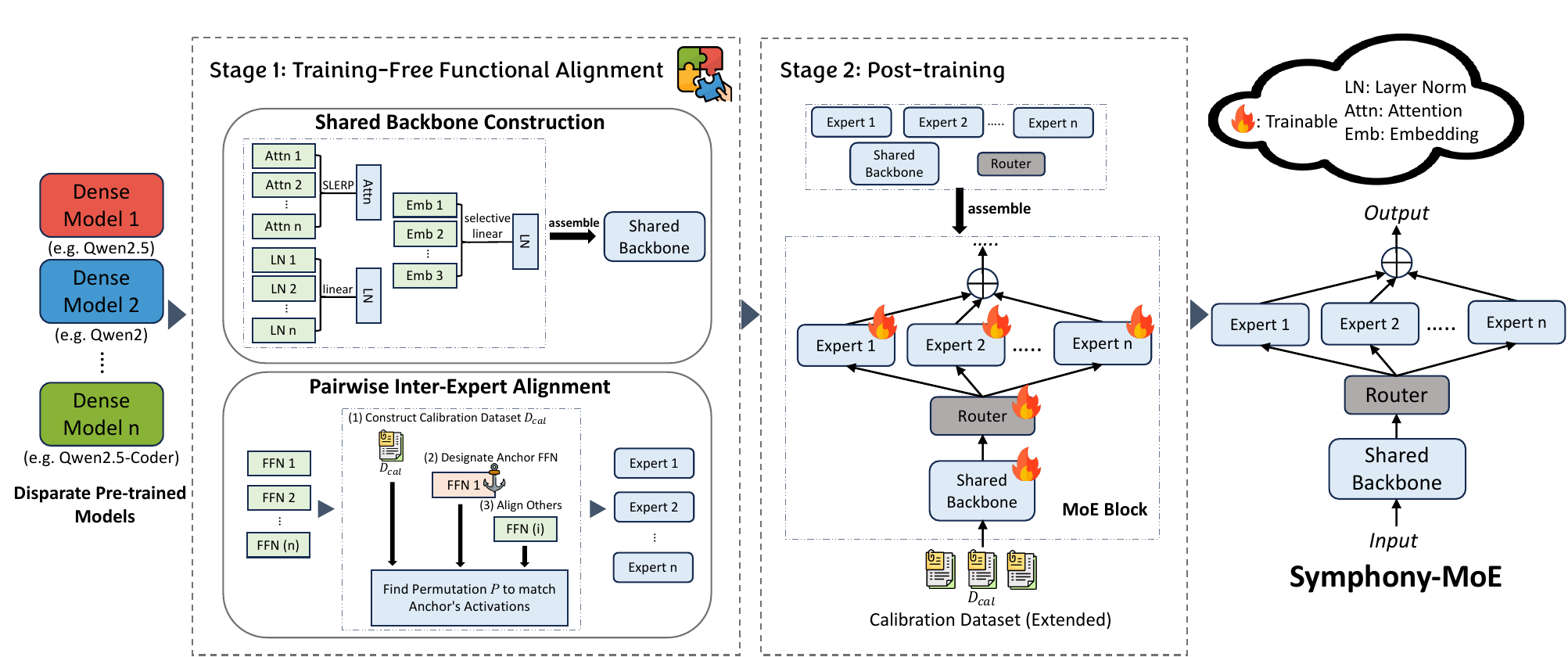} 
\caption{Overview of the Symphony-MoE construction pipeline. In Stage 1, disparate source models are harmonized without training: non-FFN layers are merged into a shared backbone using techniques such as SLERP, while FFN layers are aligned through activation-based neuron permutation. In Stage 2, the experts, router, and shared backbone undergo post-training to enable coordination among the now-compatible components.}
\label{fig_overview}
\end{figure*}

Thus, this paper addresses a core research question: \textbf{Can we effectively assemble MoE models by upcycling experts from disparately initialized source models?} Naive upcycling encounters severe \textit{parameter space misalignment}~\citep{ainsworth2022git}. Models trained with divergent histories tend to occupy incompatible regions of both the numerical and semantic parameter space. Simply concatenating such models leads to catastrophic interference, as the misaligned experts disrupt each other's internal representations. As a result, the router can no longer perform meaningful dispatching across functionally incoherent experts.

We propose Symphony-MoE, a framework for \textit{harmonizing} a collection of disparate models into a single, powerful MoE. We achieve this \textit{harmonization} through a two-stage process, where the \textit{parameter space misalignment} is resolved via training-free functional alignment.  In this first stage, we construct a coherent shared backbone using layer-aware fusion. We apply Spherical Linear Interpolation (SLERP)~\citep{wortsman2022model} to self-attention modules to maintain their geometric integrity, while using a selective linear strategy for token embeddings to handle vocabulary mismatches and simple averaging for the statistically-driven LayerNorm parameters. Crucially, to harmonize the experts themselves, we functionally align them to a shared representational space. This is achieved not by averaging, but by a precise reordering of neurons based on activation similarities, solved via a Linear Assignment Problem. This training-free process yields a pool of diverse yet functionally compatible experts, setting the stage for efficient coordination.

The second stage performs global coordination through post-training. To prevent expert collapse and encourage balanced expert utilization, this stage introduces a load-balancing loss that promotes uniform routing across experts.

The core contributions of this work are threefold:

(1) We investigate the problem of upcycling multiple identically architected yet independently pretrained models into a unified MoE model. We identify \textit{parameter space misalignment} as a key barrier to this process.

(2) We introduce a harmonization framework that addresses expert incompatibility, combining layer-aware fusion and activation-driven functional alignment to preserve the specialized knowledge embedded in each source model.

(3) Through comprehensive experiments on in-distribution and out-of-distribution datasets, we show that our method achieves strong performance and generalizes well to out-of-distribution settings, demonstrating the feasibility of harmonizing disparate pre-trained models into a coherent MoE.

\section{2 Related Works}
\subsection{2.1 Mixture-of-Experts Models}

The Mixture-of-Experts (MoE) paradigm is originally proposed as a ``divide and conquer" modular learning system~\citep{jacobs1991adaptive}. Its modern renaissance was triggered by the \textit{Sparsely-Gated MoE}~\citep{shazeer2017outrageously}, which introduced sparse, top-k routing to enable conditional computation at an unprecedented scale. This decouples the model's parameter count from its computational cost.

Subsequent advances, most notably the Switch Transformer~\citep{fedus2022switch}, simplified the routing mechanism to top-1 selection, improving training stability and reducing communication overhead. This streamlined approach established sparsity as a core strategy for scaling, as adopted in models like GLaM~\citep{du2022glam} and Mixtral~\citep{jiang2024mixtral}. However, training these massive MoE models from scratch remains prohibitively expensive, which motivates more resource-efficient construction methods.

\subsection{2.2 MoE Construction via Upcycling}

To mitigate the immense cost of training MoEs from scratch, the upcycling paradigm has emerged as a practical alternative. The vanilla approach, or Sparse Upcycling \citep{komatsuzaki2022sparse, he2024upcycling, zhao2024texttt}, involves converting a pre-trained dense model into an MoE by replicating its feed-forward network (FFN) layers to form multiple, initially identical experts. While this provides a strong initialization, it suffers from a knowledge bottleneck and a lack of initial expert diversity, as all experts originate from a single source.

To address this, Drop-Upcycling~\citep{nakamura2025drop} performs partial re-initialization to enhance expert diversity. ``Branch-Train-Mix" strategy~\citep{sukhbaatar2024branch, horoi2025less, kang2024self, li2024stitchfusion} constructs an MoE from multiple, pre-specialized dense models to inject diverse knowledge from the outset. For example, BAM~\citep{zhang2024bam} first fine-tunes copies of a base model on different domains and then uses their specialized FFN and attention layers as the experts in the final MoE. However, it introduces a new critical challenge: how to effectively fuse the knowledge from these disparate trained specialists, whose parameters reside in different regions of the loss landscape. Current methods like BAM and Self-MoE~\citep{kang2024self} rely on rudimentary fusion techniques, such as simple averaging for non-expert layers, which are not guaranteed to be optimal.

\subsection{2.3 Model Merging}

The challenge of combining disparate pre-trained models is the central focus of the model merging field~\citep{tao2024task, imfeld2023transformer}. To merge truly ``stranger" models,  advanced techniques explicitly resolve misalignment. Model Soups~\citep{wortsman2022model} averages the weights of models fine-tuned from the same checkpoint, assuming they remain in the same well-connected loss basin. Methods like PLeaS~\citep{ito2024analysis} and Git Re-Basin~\citep{ainsworth2022git} find an optimal permutation matrix to align neurons layer-wise before averaging. TIES-Merging~\citep{yadav2023ties} resolves interference between task vectors by trimming redundant updates and electing a dominant sign for parameter changes. These methods, however, are general-purpose algorithms designed to produce a single, dense consensus model. They are not tailored for the specific architectural needs of MoE construction, such as preserving expert diversity or informing the initialization of the router.

Our work bridges this gap. We integrate permutation-based alignment into the MoE construction pipeline, repurposing these techniques for a new goal. We use alignment to make disparate experts functionally compatible. This ensures they can operate within a shared coordinate system while preserving their individual specializations.

\section{3 Methods}

We introduce a two-stage framework to effectively assemble a MoE model from disparate, pre-trained source models. The first stage operates entirely without training, resolving the critical \textit{parameter space misalignment} problem through training-free functional alignment. The second stage then coordinates the now-compatible experts, unlocking their collective capabilities.

\subsection{3.1 Preliminaries and Problem Formulation}

We first establish the requisite notation for Transformer and MoE architectures and then formally define the problem.

A standard Transformer model is composed of stacked layers, each typically containing a multi-head self-attention mechanism followed by a position-wise feed-forward network (FFN).  In an MoE architecture, the dense FFN layer is replaced by a set of $N$ independent ``expert" networks $\{E_1, E_2, \ldots, E_N\}$, each with the same architecture as the original FFN. A trainable router network, $R$, directs the input tokens to a sparse subset of these experts. For each input token $x$, the router computes gating values $g(x)$ that determine how the outputs of the experts are combined:
\begin{equation}
M_{MoE}(x) = \sum_{i=1}^{N} g_i(x) \cdot E_i(x)
\end{equation}

The gating values are typically produced by a linear layer followed by a softmax function, $g(x) = Softmax(xW_g)$, where $W_g$ is a trainable routing matrix. For sparsity, often only the top-k experts with the highest gating values are activated for any given token.

Our work addresses the novel challenge of constructing an MoE model from multiple, disparate pre-trained source models. Formally, we are given a set of $N$ source models, $\{M_1, M_2, \ldots, M_N\}$. These models share an identical architecture but possess distinct sets of parameters $\{\theta_1, \theta_2, \ldots, \theta_N\}$ as a result of being trained on different datasets $\{D_1, D_2,\ldots, D_N\}$, for different tasks, or with different optimization objectives. The goal is to construct a single, powerful MoE model, $M_{MoE}$, by leveraging the FFN layers of the source models as its experts. The fundamental barrier to this goal is the severe \textbf{parameter space misalignment} between the models, where the parameters $\theta_i$ and $\theta_j$ (for $i \neq j$) occupy incompatible semantic spaces.

\subsection{3.2 Stage 1: Training-Free Functional Alignment}

Our first stage assembles a MoE model from dense models entirely without training. This process involves two key steps: fusing the shared architectural backbone and functionally aligning the FFN expert layers.

\textbf{Shared Backbone Construction.} We first construct the shared, non-expert layers. Our approach to this fusion is explicitly layer-aware. Rather than applying a single, uniform merging technique, we tailor the fusion strategy to the specific architectural function of each component. Specifically:

\textbf{$\cdot$} For the \textit{token embedding} layers, we follow the selective linear strategy implemented in MergeKit \cite{goddard2024arcee}. For tokens shared across models, the algorithm retrieves their corresponding embedding vectors from each model and performs standard linear averaging. For tokens that are unique to a specific model—i.e., not present in the vocabularies of others—the original embedding is preserved without modification.

\textbf{$\cdot$} For \textit{self-attention} modules, which are critical for contextual understanding, we apply Spherical Linear Interpolation (SLERP)~\citep{wortsman2022model}. This method better preserves the geometric integrity of the weight space and mitigates the functional degradation often caused by naive linear averaging. We apply it to all weight matrices (Q, K, V, O projections).

\textbf{$\cdot$} In contrast, the primary statistical function of \textit{LayerNorm} parameters makes a simple linear average a sufficient and stable choice ~\citep{jin2022dataless}.

\textbf{Pairwise Inter-Expert Alignment.} We designate one model's FFNs as the ``anchor" and then permute the neurons of the other models' FFNs to match the anchor's functional behavior based on activation patterns.

For each non-anchor model $M_i$ and each layer $l$, the alignment process is as follows:

1. \textit{Activation Collection}: We construct a small and diverse calibration set $D_{\mathrm{cal}}$ by sampling \textbf{equally} from each continue pre-training dataset ${D_1, D_2, \ldots, D_N}$ used in dense model training. Each instruction in $D_{\mathrm{cal}}$ is passed through both the anchor model $M_1$ and a target model $M_i$ to extract their FFN activation matrices at layer $l$, denoted as $A_1^{(l)}$ and $A_i^{(l)}$, respectively. We sample activations using a tokenizer with a merged vocabulary, extract the post-nonlinearity activation values, and perform normalization.

2.  \textit{Permutation Matching}: We find the optimal permutation matrix $P_i^{(l)}$ that aligns the neurons of $M_i$ to $M_1$ by solving the linear assignment problem:
\begin{equation}
    \min_{P \in \mathcal{P}} ||A_1^{(l)} - A_i^{(l)}P||_F^2    
\end{equation}
    
where $\mathcal{P}$ is the set of permutation matrices and $||\cdot||_F$ is the Frobenius norm. This problem can be solved efficiently with the Hungarian algorithm~\citep{mills2007dynamic}.

3. \textit{Weight Remapping}: We apply the computed permutation $P_i^{(l)}$ to the FFN weights~\citep{nasery2025pleas}. This involves permuting the output dimension of the first linear layer's weights $W_{up}$, $W_{gate}$, and the input dimension of the second linear layer's $W_{down}$ to maintain harmonization:
\begin{equation}
    W'_{up,i} = W_{up,i}^{(l)} P_i^{(l)}, \quad W'_{gate,i} = W_{gate,i}^{(l)} P_i^{(l)} 
\end{equation}
\begin{equation}
W'_{down,i} = (P_i^{(l)})^T W_{down,i}^{(l)}
\end{equation}
This procedure results in $N$ functionally aligned experts for each layer, all operating in a consistent parameter space. 

We analyzed the computational cost to verify the scalability. The time complexity of this solution is polynomial. The analysis can be found in the Appendix.

\subsection{3.3 Stage 2: Activating the MoE via  Post-training}

In Stage 2, we integrate the constructed shared backbone with the expert models and introduce a randomly initialized router, implemented as a simple linear layer. The calibration dataset $D_{\mathrm{cal}}$, introduced in Stage 1, is extended, enabling the MoE to learn how to coordinate the experts.  The expert layers, shared backbone, and the router are trained on the extended calibration dataset $D_{cal}$.

We adopt a \textit{top-2 routing} mechanism and train the MoE model using a composite loss function:
\begin{equation}
\mathcal{L}_{total} = \mathcal{L}_{lm} + \lambda \cdot \mathcal{L}_{bal}
\end{equation}
where $\mathcal{L}_{lm}$ is the standard causal language modeling objective, and $\mathcal{L}_{bal}$ is a load-balancing regularization term introduced in Switch Transformer~\cite{fedus2022switch}. The coefficient $\lambda$ is fixed at 0.01 in all experiments. 

Notably, the router is not provided with explicit supervision indicating which expert to select for each token. It learns expert assignment implicitly through continual pre-training on $D_{\mathrm{cal}}$.

\section{4 Experiments}
In this section, we aim to address the following key research questions: (1) How does Symphony-MoE perform at different model scales or types, compared to baseline models on both in-distribution and out-of-distribution (OOD) datasets? (2) Does our proposed alignment method effectively mitigate parameter space misalignment across experts? (3) How does the choice of anchor model influence the final performance of Symphony-MoE? (4) What is the impact of varying the number of experts on model performance?  (5) Ablation Analysis: How sensitive is the system to variations in alignment and merging strategies?

\subsection{4.1 Experimental Setup}
\subsubsection{4.1.1 Build Disparately Initialized Dense Models}
\begin{table*}[t] 
\centering
\begin{tabular}{llcccccccc}
\toprule
 &  & &  &  \multicolumn{2}{c}{ID}&   && &  OOD\\
&  \vrule & MMLU& GSM8K& BBH & HumanEval & TruthfulQA & \textbf{Avg.*}& \vrule & MedCQA\\ 
\midrule
\textit{Dense (0.5B)}&   &  &  &  &  &   &&  &   \\
General ($M_1$, ANC)& \vrule & \underline{44.27}$_{\pm 0.40}$& 18.87$_{\pm 1.07}$ & 27.46$_{\pm 0.50}$ & 24.40$_{\pm 3.36}$ & 27.29$_{\pm 1.56}$ & 28.46$^*$ &  \vrule & 26.20$_{\pm 0.19}$\\ 
Math ($M_2$)& \vrule & 42.07$_{\pm 0.40}$ & 19.33$_{\pm 1.08}$ &  22.72$_{\pm 0.46}$  & 20.73$_{\pm 3.17}$ & 28.40$_{\pm 1.58}$ & 26.65$^*$  &\vrule  & 26.23$_{\pm 0.19}$\\
Code ($M_3$)& \vrule & 30.96$_{\pm 0.38}$ & 20.48$_{\pm 0.41}$ &  25.33$_{\pm 0.49}$ & \textbf{28.71}$_{\pm 3.27}$& 25.81$_{\pm 1.55}$ & 26.26$^*$ & \vrule & 25.93$_{\pm 0.19}$\\ 
Science ($M_4$)&  \vrule & 39.25$_{\pm 0.41}$ & 21.30$_{\pm 1.12}$ & 22.98$_{\pm 0.48}$ & 18.29$_{\pm 3.02}$ & 30.01$_{\pm 1.50}$ & 26.37$^*$ & \vrule & 25.56$_{\pm 0.19}$\\ 
\midrule 
\textit{ MoE (0.5B $\times$ 4)}&   &  &  &  &  &   &&   &   \\
BTX& \vrule & 37.28$_{\pm 0.61}$& 18.95$_{\pm 0.42}$&25.57$_{\pm 0.48}$& 26.59$_{\pm 3.15}$&  23.90$_{\pm 1.19}$&26.46$^*$& \vrule & 24.02$_{\pm 0.19}$\\
BAM& \vrule & 42.76$_{\pm 0.70}$& 19.15$_{\pm 0.68}$&\underline{27.68}$_{\pm 0.58}$& 26.77$_{\pm 3.07}$&  26.54$_{\pm 1.61}$&28.58$^*$&\vrule & 26.79$_{\pm 0.19}$\\
Drop& \vrule & 44.08$_{\pm 0.40}$& \underline{22.85}$_{\pm 0.39}$&26.51$_{\pm 0.48}$& 25.52$_{\pm 3.19}$&  \underline{30.81}$_{\pm 1.55}$&\underline{29.95}$^*$&\vrule & \underline{27.95}$_{\pm 0.20}$\\
Symphony (Ours)& \vrule & \textbf{45.10}$_{\pm 0.40}$& \textbf{24.57}$_{\pm 1.11}$&\textbf{29.64}$_{\pm 0.48}$& \underline{28.02}$_{\pm 3.12}$&  \textbf{31.54}$_{\pm 1.50}$&\textbf{31.77}$^*$& \vrule & \textbf{29.07}$_{\pm 0.19}$\\
\bottomrule
\end{tabular}
\caption{Performance comparison of dense (\textbf{Qwen2/2.5 0.5B}) and MoE models (upcycled from dense models) on in-distribution (ID) and out-of-distribution (OOD) data.  The best and second-best results are \textbf{bolded} and \underline{underlined}. ``ANC" stands for anchor model. ``Avg*" stands for the average scores of ID datasets.}
\label{tab:main_results_0.5B}
\end{table*}

\begin{table*}[h] 
\centering
\begin{tabular}{llcccccccc}
\toprule
 &  & &  &  \multicolumn{2}{c}{ID}&   &  &  &  OOD \\
 &  \vrule & MMLU& GSM8K & BBH& HumanEval & TruthfulQA & \textbf{Avg.$^*$}& \vrule & MedCQA \\ 
\midrule
 \textit{Dense (1.5B)}&   &  &  &  &  &   & &   &   \\
 General ($M_1$, ANC)& \vrule & \textbf{59.57}$_{\pm 0.39}$ & 33.72$_{\pm 0.39}$ & 44.39$_{\pm 0.55}$ & 37.19$_{\pm 3.79}$ & 29.87$_{\pm 1.60}$ & 40.95$^*$ &  \vrule & 30.39$_{\pm 0.20}$ \\ 
Math ($M_2$)& \vrule & 56.73$_{\pm 0.39}$ &  34.11$_{\pm 0.50}$  & 43.97$_{\pm 0.54}$  & 37.20$_{\pm 3.79}$ & 31.33$_{\pm 1.62}$  &  40.67$^*$  &\vrule  & 30.36$_{\pm 0.20}$ \\
Code ($M_3$)& \vrule & 43.74$_{\pm 0.41}$ & 35.86$_{\pm 1.33}$  & 36.17$_{\pm 0.55}$ & \textbf{43.41}$_{\pm 3.81}$ & 26.19$_{\pm 1.51}$ & 37.07$^*$ & \vrule & 28.72$_{\pm 0.19}$\\ 
Science ($M_4$)&  \vrule & 54.60$_{\pm 0.40}$ & 32.45$_{\pm 1.30}$ & 36.05$_{\pm 0.53}$ & 35.37$_{\pm 3.74}$ & 30.78$_{\pm 1.57}$ & 37.85$^*$ & \vrule &  29.72$_{\pm 0.19}$ \\ 
\midrule 
 \textit{MoE (1.5B $\times$ 4)}&  &  &  &  &  &   &  &   &   \\
BTX& \vrule  & 45.12$_{\pm 0.88}$  & 30.35$_{\pm 0.45}$ & 40.11$_{\pm 0.58}$  &  29.08$_{\pm 3.15}$ & 25.02$_{\pm 1.15}$  &  33.94$^*$ & \vrule & 26.92$_{\pm 0.18}$ \\
    BAM& \vrule & 50.77$_{\pm 0.75}$ & \underline{36.99}$_{\pm 0.61}$ & \underline{45.01}$_{\pm 0.59}$ & 37.84$_{\pm 2.89}$ & 27.14$_{\pm 1.89}$ & 39.55$^*$ &\vrule & 28.97$_{\pm 0.18}$\\
Drop& \vrule  & 57.14$_{\pm 0.81}$  &  34.92$_{\pm 1.12}$  & 44.83 $_{\pm 0.43}$  & 36.72$_{\pm 1.89}$ & \underline{31.88}$_{\pm 1.49}$  & \underline{41.10}$^*$ &\vrule & \underline{32.90}$_{\pm 0.19}$ \\
Symphony (Ours)& \vrule & \underline{58.91}$_{\pm 0.31}$ & \textbf{39.12}$_{\pm 1.09}$ & \textbf{46.97}$_{\pm 0.36}$ & \underline{42.39}$_{\pm 3.69}$ & \textbf{32.95}$_{\pm 2.31}$ & \textbf{44.07}$^*$ & \vrule & \textbf{35.26}$_{\pm 0.22}$\\
\bottomrule
\end{tabular}
\caption{Performance comparison of dense (\textbf{Qwen2/2.5 1.5B}) and MoE models (upcycled from dense models) on in-distribution (ID) and out-of-distribution (OOD) data.  The best and second-best results are \textbf{bolded} and \underline{underlined}. ``ANC" stands for anchor model. ``Avg*" stands for the average scores of ID datasets.}
\label{tab:main_results_1.5B}
\end{table*}

To ensure a rigorous evaluation of our proposed merging method, we first constructed a set of four disparately initialized dense models at scale of 0.5/1.5B. This experimental suite was designed to reflect real-world disparity by systematically incorporating variations across three dimensions: model versions, pre-training data domains, and downstream instruction-tuning tasks.

Our framework is built upon three distinct foundation checkpoints, each contributing a different source of disparity.  The first is Qwen2.5-Base, a general-purpose model trained on broad-domain natural language data, with a focus on tasks involving understanding and generation. To introduce task- and data-level disparity, we include Qwen2.5-Coder~\citep{hui2024qwen2}, which is primarily trained on code-related corpora and optimized for tasks such as code completion and code–natural language interconversion. Finally, to incorporate version-level disparity, we use Qwen2-Base~\citep{team2024qwen2}, an architecturally compatible predecessor of Qwen2.5-Base with a completely different pre-training history.

Based on the three foundational models, we derived four specialist experts through instruction fine-tuning. The generalist expert ($M_1$) and the mathematics expert ($M_2$) were obtained by fine-tuning Qwen2.5-Base on the Alpaca~\citep{maeng2017alpaca} and MetaMathQA~\citep{yu2024metamath} datasets, respectively.
The code expert ($M_3$) was derived from our code-specialized foundation using the CodeAlpaca~\citep{li2024instructcoder} dataset. 
The science expert ($M_4$) was created by fine-tuning Qwen2-Base on the SciQAG~\citep{wan2024sciqag} dataset.

Further training details are provided in the Appendix.

\subsubsection{4.1.2 Evaluation Datasets and Metrics}
We evaluate our method on five in-distribution datasets: MMLU (General) ~\citep{hendrycks2021measuringmassivemultitasklanguage}, GSM8K (Math)~\citep{cobbe2021trainingverifierssolvemath}, BBH (General, Reasoning) ~\citep{suzgun2022challengingbigbenchtaskschainofthought}, HumanEval (Code)~\citep{chen2021evaluating}, and TruthfulQA (Science)~\citep{lin2022truthfulqameasuringmodelsmimic}. We also include an out-of-distribution benchmark, MedCQA (Medicine)~\citep{shoham2024medconceptsqa}.

For evaluation, we follow the standard metrics used in each benchmark: MMLU and MedCQA are evaluated using zero-shot accuracy, GSM8K and BBH use eight-shot accuracy, HumanEval is measured with pass@1, and TruthfulQA adopts the MC1 metric.

\subsubsection{4.1.3 Baselines}
We compare our upcycling approach against two categories: the dense models $M_1$, $M_2$, $M_3$, $M_4$ that we upcycle, and recent upcycling methods (BTX \cite{sukhbaatar2024branch}, BAM \cite{zhang2024bam}, Drop \cite{nakamura2025drop}). To ensure fairness, the baseline method uses the same data as ours for fine-tuning the dense model. We reproduced the results according to the original paper, and the reproduction details are in the appendix.

\subsection{4.2 Implementation Details}
\begin{figure*}[t]
\centering
\includegraphics[width=0.98\textwidth]{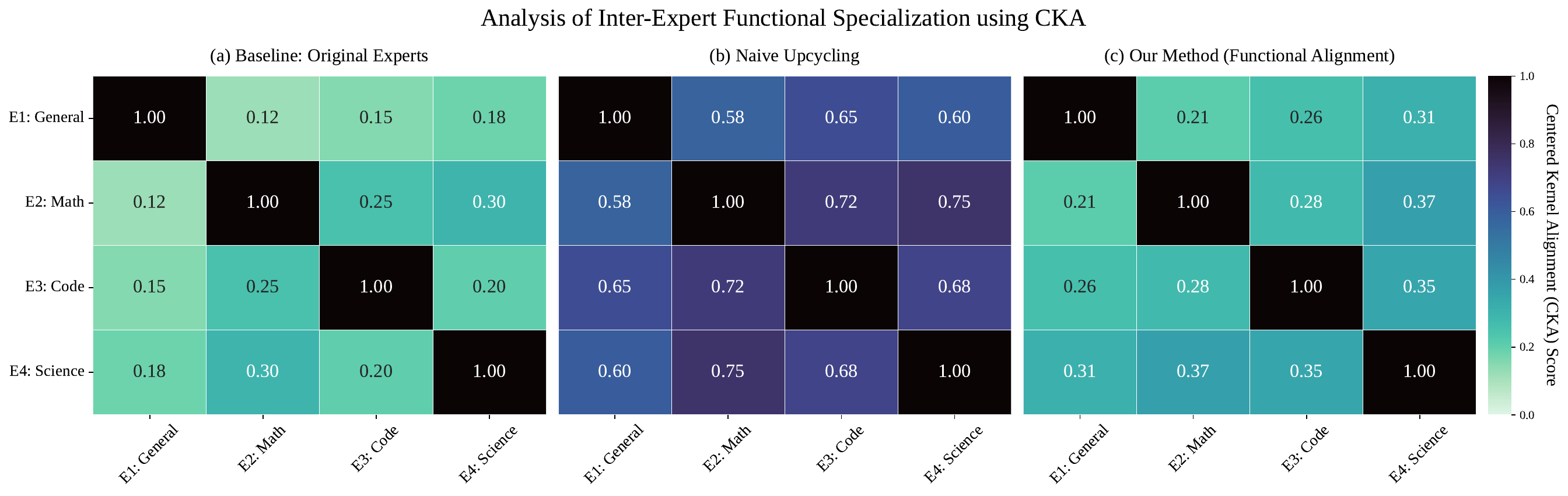} 
\caption{Quantitative analysis of inter-expert functional specialization using Centered Kernel Alignment (CKA). \textbf{Lower CKA} scores indicate greater \textbf{functional specialization}. \textbf{High CKA} scores between experts reflect \textbf{parameter space misalignment}, caused by failing to \textit{align neurons functionally} during merging. This misalignment leads to representational collapse, erasing the distinct capabilities of individual experts.}

\label{fig_cka}
\end{figure*}

Our experiments were conducted on 24 NVIDIA V100 GPUs. The dense model was implemented using LLaMA-Factory~\citep{zheng2024llamafactory}, and trained for 2 epochs. 

In Stage 1: Alignment, we randomly sampled from each training dataset to construct the calibration set $D_{\mathrm{cal}}$, resulting in a total of \textbf{10.4M tokens}. The training data is sampled from SlimPajama\footnote{https://huggingface.co/datasets/cerebras/SlimPajama-627B} (General, Code), 
Finemath\footnote{\url{https://huggingface.co/datasets/HuggingFaceTB/finemath}}, 
and Scientific Papers\footnote{\url{https://huggingface.co/datasets/scientific_papers}}. To ensure consistent alignment and integration across expert models, we fixed a general-purpose expert model $M_1$ as the anchor model. To ensure that the knowledge of each expert is equally integrated and to avoid introducing any bias, the weights for merging the 4 dense models are set to 0.25.

In Stage 2: Post-training, we increased the sampled data size, expanding $D_{\mathrm{cal}}$ to \textbf{5B tokens}. We set the cutoff length to 2048 and train the model for 6 epochs with a batch size of 2. The learning rate is fixed at 5e-5, and optimization is performed using AdamW~\citep{zhuang2022understanding} with a maximum gradient norm of 1.0.

In the evaluation phase, we used lm-evaluation-harness-v0.4~\citep{eval-harness} to obtain scores for datasets such as MMLU and GSM8K.

\subsection{4.3 Main Results} \label{main}

\textbf{Overall Performance.} As summarized in Table~\ref{tab:main_results_0.5B} and \ref{tab:main_results_1.5B}, Symphony-MoE achieves the highest average score across all evaluation datasets, substantially outperforming all MoE baselines, including BTX, BAM, and Drop. In contrast, individual dense models demonstrate strong domain specialization but suffer severe performance degradation on out-of-domain tasks, indicating limited generalization.

\textbf{In-Domain Expertise Preservation.} Importantly, Symphony-MoE is not a naive average of its experts. On in-domain tasks, it effectively retains the specialized capabilities of its constituent models. For example, in the HumanEval, although the dedicated code expert $M_3$ achieves the top score, Symphony-MoE performs comparably, trailing by only 1–2 percentage points.

\textbf{Out-of-Distribution Generalization.} Beyond domain retention, Symphony-MoE demonstrates superior generalization ability, outperforming all baselines on MedCQA. This suggests that our activation-based alignment strategy enables the model to internalize more transferable and fundamental reasoning patterns, rather than merely interpolating between expert skills.

\textbf{Robustness Across Scales.} We further validate these findings across two model scales—4 experts × 0.5B and 4 experts × 1.5B. In both settings, Symphony-MoE consistently exhibits strong average performance, competitive in-domain expertise, and leading out-of-distribution generalization, confirming the robustness of our approach.

\textbf{Robustness Across Types.} We further validated the effectiveness of Symphony-MoE using the Llama model as the source architecture. As shown in the Appendix, our method yields consistent improvements, demonstrating its robustness across different backbone model types.

\subsection{4.4 Analysis of Inter-Expert Functional Specialization} \label{specialization}
To quantitatively assess the effectiveness of our functional alignment method, we compute the pairwise functional specialization between expert layers using Centered Kernel Alignment (CKA)~\citep{cortes2012algorithms}, a widely adopted technique for comparing neural network representations. A lower CKA score indicates greater functional specialization. We evaluate three scenarios: (a) the original, unmerged expert models as a reference baseline; (b) a naive merging approach without alignment; (c) our method employing activation-based functional alignment.

As shown in Figure~\ref{fig_cka}, the naive merging strategy leads to a severe representational collapse. Experts trained independently develop distinct internal coordinate systems; merging them without alignment forces the model to conflate functionally unrelated neurons. This results in inflated inter-expert CKA scores (e.g., 0.65–0.75), signaling that \textit{unique functional fingerprints of the experts have been blurred into a redundant subspace}. In such a scenario, the router lacks the discriminative basis to perform specialized dispatching.

In contrast, our method explicitly resolves this misalignment by permuting neurons to form a functionally coherent coordinate system prior to merging. Figure~\ref{fig_cka}(c) demonstrates that this alignment dramatically restores inter-expert specialization, with CKA scores returning to near-optimal levels comparable to those of the original experts (Figure~\ref{fig_cka}(a)). This provides empirical evidence that our harmonization framework effectively mitigates parameter space misalignment.

\subsection{4.5 Analysis of Anchor Model Selection} \label{anchor}
\begin{figure}[t]
\centering
\includegraphics[width=0.47\textwidth]{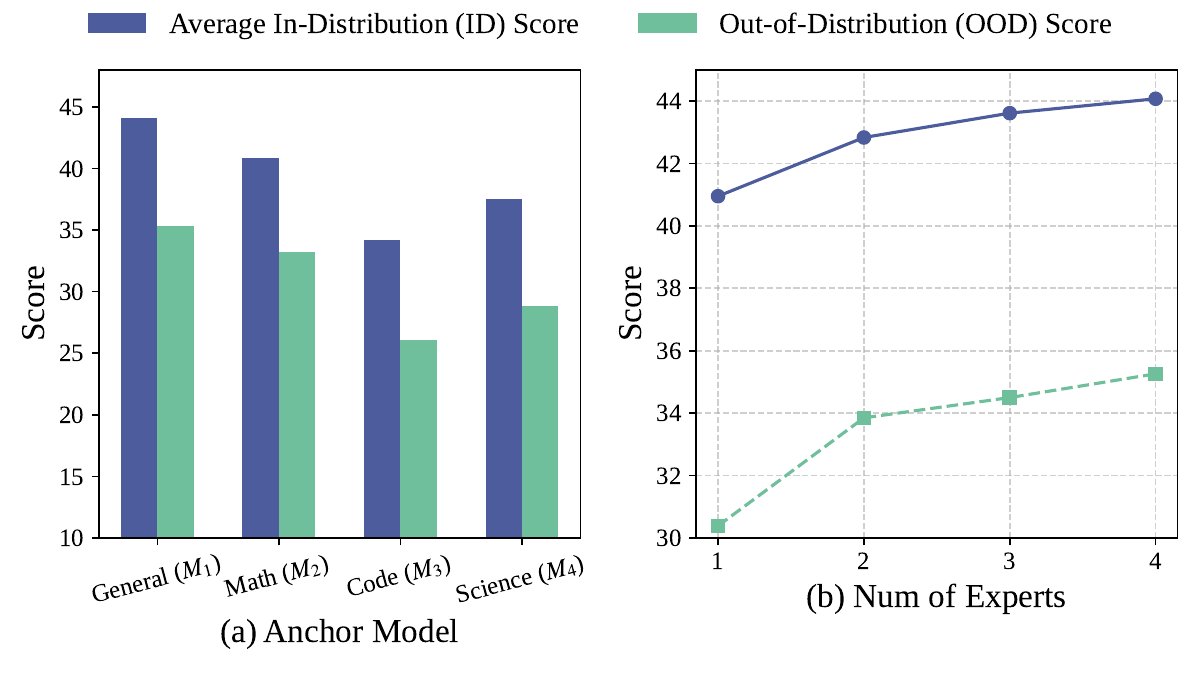} 
\caption{Performance analysis of Symphony-MoE. (a) Average in-distribution (ID) and out-of-distribution (OOD) scores under different anchor model choices. (b) Impact of increasing the number of experts from 1 to 4 on ID and OOD performance.}
\label{fig_anchor}
\end{figure}

To isolate the impact of anchor model selection, we fixed all other experimental conditions and varied only the anchor model used in the first stage. Specifically, we constructed four Symphony-MoE (1.5B$\times$4) models using $M_1$, $M_2$, $M_3$, and $M_4$ as anchors, respectively, and evaluated their average performance across both in-distribution (ID) and out-of-distribution (OOD) datasets.

As shown in the Figure~\ref{fig_anchor}(a), using the general-purpose expert model $M_1$ as the anchor yields the highest average scores on both ID and OOD tasks. This suggests that a balanced and non-specialized functional space plays a critical role in constructing an effective “omnipotent” MoE. It facilitates collaboration among experts without constraining them to overly narrow subspaces.

\subsection{4.6 Analysis of the Number of Experts} \label{noe}
In Figure~\ref{fig_anchor}(b), we illustrate the effect of varying the number of experts in Symphony-MoE (1.5B$\times$4). Specifically, experts are added incrementally in a fixed order: $M_1$ alone for one expert; $M_1$ and $M_2$ for two; $M_1$, $M_2$, and $M_3$ for three; and all four experts for the complete model. The results show that increasing the number of experts consistently improves performance on both ID and OOD tasks, with the addition of the mathematics expert ($M_2$) yielding the most substantial performance gain.

\subsection{4.7 Ablation Analysis} \label{ablation}

\begin{table}[t]
\centering
\begin{tabular}{lcc}
\toprule
\textbf{Model / Configuration	}& \textbf{ID Avg.	}& \textbf{OOD}\\ 
\midrule
(+) Full Method& 44.07& 35.26\\ 
(-) No Functional Alignment	& 33.94& 26.92\\ 
 \quad   (-) Align 80\% of the Neurons  & 43.57 & 34.61 \\
(-) Naive Attention Merge	& 37.28& 31.77\\ 
(-) Naive Embedding Merge 	& 40.19& 31.95\\ 
(-) Biased Calibration Data	& 39.70& 29.44\\ 
\bottomrule
\end{tabular}
\caption{Ablation analysis of key components in Symphony-MoE (1.5B x 4).}
\label{tab:ablation}
\end{table}

We assessed the impact of key components on the performance of the Symphony-MoE model (1.5B × 4 configuration) through an ablation study, where each core module was individually removed or replaced. The results are presented in Table~\ref{tab:ablation}. Specifically, \textit{Naive Attention Merge} replaces the SLERP fusion strategy with simple linear averaging when constructing the shared backbone for self-attention layers in Stage 1. This setting quantifies the contribution of SLERP to preserving the functional integrity of attention mechanisms. \textit{Naive Embedding Merge} replaces MergeKit with a naive linear averaging of overlapping token embeddings during the construction of the shared embedding layer. \textit{Biased Calibration Data} alters the sampling strategy of the calibration dataset $D_{\mathrm{cal}}$ in the alignment stage, using samples exclusively from the training set of the general expert $M_1$ rather than from all expert domains.

The results highlight the critical role of functional alignment: removing it causes catastrophic performance collapse, confirming that resolving parameter space inconsistencies is fundamental to the framework. Aligning only 80\% of the neurons did not result in a significant decrease in the model's performance, demonstrating the robustness of the alignment method.
Replacing SLERP with linear averaging leads to substantial degradation, supporting our hypothesis that the geometry of attention weights is essential for preserving functionality. While simplifying the embedding fusion yields only a moderate drop, it nonetheless indicates the value of layer-aware merging in the shared backbone. Notably, using biased calibration data results in the most severe performance decline after removing alignment, underscoring the importance of domain diversity in constructing $D_{\mathrm{cal}}$ for effective inter-expert coordination.

\section{5 Conclusion}

This paper presents Symphony-MoE, a framework for building powerful MoE models by upcycling experts from multiple, diverse pre-trained sources. We tackle the core challenge of parameter space misalignment through a two-stage process. Experiments show that Symphony-MoE outperforms baselines on multi-domain tasks and generalizes well to out-of-distribution data. This approach offers a scalable path to leverage the collective knowledge embedded in existing specialized models.

\textbf{Limitations.} Our current framework requires that all source models share an identical architecture and are all language models. This constraint limits the pool of potential experts that can be integrated. Future work could focus on developing more advanced alignment techniques capable of harmonizing models with \textit{minor architectural dissimilarities} or \textit{modality differences}. The effectiveness of functional alignment depends on the diversity of the calibration dataset. Although we propose a straightforward sampling strategy, optimizing the process to ensure a highly diverse set of experts presents a significant challenge.

\bibliography{aaai2026}

\begin{thebibliography}{43}
\providecommand{\natexlab}[1]{#1}

\bibitem[{Ainsworth, Hayase, and Srinivasa(2022)}]{ainsworth2022git}
Ainsworth, S.~K.; Hayase, J.; and Srinivasa, S. 2022.
\newblock Git re-basin: Merging models modulo permutation symmetries.
\newblock \emph{arXiv preprint arXiv:2209.04836}.

\bibitem[{Chen(2021)}]{chen2021evaluating}
Chen, M. 2021.
\newblock Evaluating large language models trained on code.
\newblock \emph{arXiv preprint arXiv:2107.03374}.

\bibitem[{Cobbe et~al.(2021)Cobbe, Kosaraju, Bavarian, Chen, Jun, Kaiser, Plappert, Tworek, Hilton, Nakano, Hesse, and Schulman}]{cobbe2021trainingverifierssolvemath}
Cobbe, K.; Kosaraju, V.; Bavarian, M.; Chen, M.; Jun, H.; Kaiser, L.; Plappert, M.; Tworek, J.; Hilton, J.; Nakano, R.; Hesse, C.; and Schulman, J. 2021.
\newblock Training Verifiers to Solve Math Word Problems.
\newblock arXiv:2110.14168.

\bibitem[{Cortes, Mohri, and Rostamizadeh(2012)}]{cortes2012algorithms}
Cortes, C.; Mohri, M.; and Rostamizadeh, A. 2012.
\newblock Algorithms for learning kernels based on centered alignment.
\newblock \emph{The Journal of Machine Learning Research}, 13(1): 795--828.

\bibitem[{Du et~al.(2022)Du, Huang, Dai, Tong, Lepikhin, Xu, Krikun, Zhou, Yu, Firat et~al.}]{du2022glam}
Du, N.; Huang, Y.; Dai, A.~M.; Tong, S.; Lepikhin, D.; Xu, Y.; Krikun, M.; Zhou, Y.; Yu, A.~W.; Firat, O.; et~al. 2022.
\newblock Glam: Efficient scaling of language models with mixture-of-experts.
\newblock In \emph{International conference on machine learning}, 5547--5569. PMLR.

\bibitem[{Fedus, Zoph, and Shazeer(2022)}]{fedus2022switch}
Fedus, W.; Zoph, B.; and Shazeer, N. 2022.
\newblock Switch transformers: Scaling to trillion parameter models with simple and efficient sparsity.
\newblock \emph{Journal of Machine Learning Research}, 23(120): 1--39.

\bibitem[{Gao et~al.(2024)Gao, Tow, Abbasi, Biderman, Black, DiPofi, Foster, Golding, Hsu, Le~Noac'h, Li, McDonell, Muennighoff, Ociepa, Phang, Reynolds, Schoelkopf, Skowron, Sutawika, Tang, Thite, Wang, Wang, and Zou}]{eval-harness}
Gao, L.; Tow, J.; Abbasi, B.; Biderman, S.; Black, S.; DiPofi, A.; Foster, C.; Golding, L.; Hsu, J.; Le~Noac'h, A.; Li, H.; McDonell, K.; Muennighoff, N.; Ociepa, C.; Phang, J.; Reynolds, L.; Schoelkopf, H.; Skowron, A.; Sutawika, L.; Tang, E.; Thite, A.; Wang, B.; Wang, K.; and Zou, A. 2024.
\newblock The Language Model Evaluation Harness.

\bibitem[{Goddard et~al.(2024)Goddard, Siriwardhana, Ehghaghi, Meyers, Karpukhin, Benedict, McQuade, and Solawetz}]{goddard2024arcee}
Goddard, C.; Siriwardhana, S.; Ehghaghi, M.; Meyers, L.; Karpukhin, V.; Benedict, B.; McQuade, M.; and Solawetz, J. 2024.
\newblock Arcee's mergekit: A toolkit for merging large language models.
\newblock \emph{arXiv preprint arXiv:2403.13257}.

\bibitem[{He et~al.(2024)He, Khattar, Prenger, Korthikanti, Yan, Liu, Fan, Aithal, Shoeybi, and Catanzaro}]{he2024upcycling}
He, E.; Khattar, A.; Prenger, R.; Korthikanti, V.; Yan, Z.; Liu, T.; Fan, S.; Aithal, A.; Shoeybi, M.; and Catanzaro, B. 2024.
\newblock Upcycling large language models into mixture of experts.
\newblock \emph{arXiv preprint arXiv:2410.07524}.

\bibitem[{Hendrycks et~al.(2021)Hendrycks, Burns, Basart, Zou, Mazeika, Song, and Steinhardt}]{hendrycks2021measuringmassivemultitasklanguage}
Hendrycks, D.; Burns, C.; Basart, S.; Zou, A.; Mazeika, M.; Song, D.; and Steinhardt, J. 2021.
\newblock Measuring Massive Multitask Language Understanding.
\newblock arXiv:2009.03300.

\bibitem[{Horoi et~al.(2025)Horoi, Wolf, Belilovsky, and Dziugaite}]{horoi2025less}
Horoi, S.; Wolf, G.; Belilovsky, E.; and Dziugaite, G.~K. 2025.
\newblock Less is More: Undertraining Experts Improves Model Upcycling.
\newblock \emph{arXiv preprint arXiv:2506.14126}.

\bibitem[{Hui et~al.(2024)Hui, Yang, Cui, Yang, Liu, Zhang, Liu, Zhang, Yu, Lu et~al.}]{hui2024qwen2}
Hui, B.; Yang, J.; Cui, Z.; Yang, J.; Liu, D.; Zhang, L.; Liu, T.; Zhang, J.; Yu, B.; Lu, K.; et~al. 2024.
\newblock Qwen2. 5-coder technical report.
\newblock \emph{arXiv preprint arXiv:2409.12186}.

\bibitem[{Imfeld et~al.(2023)Imfeld, Graldi, Giordano, Hofmann, Anagnostidis, and Singh}]{imfeld2023transformer}
Imfeld, M.; Graldi, J.; Giordano, M.; Hofmann, T.; Anagnostidis, S.; and Singh, S.~P. 2023.
\newblock Transformer fusion with optimal transport.
\newblock \emph{arXiv preprint arXiv:2310.05719}.

\bibitem[{Ito, Yamada, and Kumagai(2024)}]{ito2024analysis}
Ito, A.; Yamada, M.; and Kumagai, A. 2024.
\newblock Analysis of Linear Mode Connectivity via Permutation-Based Weight Matching: With Insights into Other Permutation Search Methods.
\newblock \emph{arXiv preprint arXiv:2402.04051}.

\bibitem[{Jacobs et~al.(1991)Jacobs, Jordan, Nowlan, and Hinton}]{jacobs1991adaptive}
Jacobs, R.~A.; Jordan, M.~I.; Nowlan, S.~J.; and Hinton, G.~E. 1991.
\newblock Adaptive mixtures of local experts.
\newblock \emph{Neural computation}, 3(1): 79--87.

\bibitem[{Jiang et~al.(2024)Jiang, Sablayrolles, Roux, Mensch, Savary, Bamford, Chaplot, Casas, Hanna, Bressand et~al.}]{jiang2024mixtral}
Jiang, A.~Q.; Sablayrolles, A.; Roux, A.; Mensch, A.; Savary, B.; Bamford, C.; Chaplot, D.~S.; Casas, D. d.~l.; Hanna, E.~B.; Bressand, F.; et~al. 2024.
\newblock Mixtral of experts.
\newblock \emph{arXiv preprint arXiv:2401.04088}.

\bibitem[{Jin et~al.(2022)Jin, Ren, Preotiuc-Pietro, and Cheng}]{jin2022dataless}
Jin, X.; Ren, X.; Preotiuc-Pietro, D.; and Cheng, P. 2022.
\newblock Dataless knowledge fusion by merging weights of language models.
\newblock \emph{arXiv preprint arXiv:2212.09849}.

\bibitem[{Kang et~al.(2024)Kang, Karlinsky, Luo, Wang, Hansen, Glass, Cox, Panda, Feris, and Ritter}]{kang2024self}
Kang, J.; Karlinsky, L.; Luo, H.; Wang, Z.; Hansen, J.; Glass, J.; Cox, D.; Panda, R.; Feris, R.; and Ritter, A. 2024.
\newblock Self-moe: Towards compositional large language models with self-specialized experts.
\newblock \emph{arXiv preprint arXiv:2406.12034}.

\bibitem[{Komatsuzaki et~al.(2022)Komatsuzaki, Puigcerver, Lee-Thorp, Ruiz, Mustafa, Ainslie, Tay, Dehghani, and Houlsby}]{komatsuzaki2022sparse}
Komatsuzaki, A.; Puigcerver, J.; Lee-Thorp, J.; Ruiz, C.~R.; Mustafa, B.; Ainslie, J.; Tay, Y.; Dehghani, M.; and Houlsby, N. 2022.
\newblock Sparse upcycling: Training mixture-of-experts from dense checkpoints.
\newblock \emph{arXiv preprint arXiv:2212.05055}.

\bibitem[{Li et~al.(2024{\natexlab{a}})Li, Zhang, Zhao, Gao, and Li}]{li2024stitchfusion}
Li, B.; Zhang, D.; Zhao, Z.; Gao, J.; and Li, X. 2024{\natexlab{a}}.
\newblock Stitchfusion: Weaving any visual modalities to enhance multimodal semantic segmentation.
\newblock \emph{arXiv preprint arXiv:2408.01343}.

\bibitem[{Li et~al.(2024{\natexlab{b}})Li, Hu, Zhao, Chen, Xie, Liu, Shieh, and He}]{li2024instructcoder}
Li, K.; Hu, Q.; Zhao, J.~X.; Chen, H.; Xie, Y.; Liu, T.; Shieh, M.; and He, J. 2024{\natexlab{b}}.
\newblock InstructCoder: Instruction Tuning Large Language Models for Code Editing.
\newblock In \emph{Proceedings of the 62nd Annual Meeting of the Association for Computational Linguistics (Volume 4: Student Research Workshop)}, 473--493.

\bibitem[{Lin, Hilton, and Evans(2022)}]{lin2022truthfulqameasuringmodelsmimic}
Lin, S.; Hilton, J.; and Evans, O. 2022.
\newblock TruthfulQA: Measuring How Models Mimic Human Falsehoods.
\newblock arXiv:2109.07958.

\bibitem[{Maeng, Colin, and Lucia(2017)}]{maeng2017alpaca}
Maeng, K.; Colin, A.; and Lucia, B. 2017.
\newblock Alpaca: Intermittent execution without checkpoints.
\newblock \emph{Proceedings of the ACM on Programming Languages}, 1(OOPSLA): 1--30.

\bibitem[{Mills-Tettey, Stentz, and Dias(2007)}]{mills2007dynamic}
Mills-Tettey, G.~A.; Stentz, A.; and Dias, M.~B. 2007.
\newblock The dynamic hungarian algorithm for the assignment problem with changing costs.
\newblock \emph{Robotics Institute, Pittsburgh, PA, Tech. Rep. CMU-RI-TR-07-27}, 7.

\bibitem[{Nakamura et~al.(2025)Nakamura, Akiba, Fujii, Oda, Yokota, and Suzuki}]{nakamura2025drop}
Nakamura, T.; Akiba, T.; Fujii, K.; Oda, Y.; Yokota, R.; and Suzuki, J. 2025.
\newblock Drop-Upcycling: Training sparse mixture of experts with partial re-initialization.
\newblock \emph{arXiv preprint arXiv:2502.19261}.

\bibitem[{Nasery et~al.(2025)Nasery, Hayase, Koh, and Oh}]{nasery2025pleas}
Nasery, A.; Hayase, J.; Koh, P.~W.; and Oh, S. 2025.
\newblock PLeaS-Merging Models with Permutations and Least Squares.
\newblock In \emph{Proceedings of the Computer Vision and Pattern Recognition Conference}, 30493--30502.

\bibitem[{Roziere et~al.(2023)Roziere, Gehring, Gloeckle, Sootla, Gat, Tan, Adi, Liu, Sauvestre, Remez et~al.}]{roziere2023code}
Roziere, B.; Gehring, J.; Gloeckle, F.; Sootla, S.; Gat, I.; Tan, X.~E.; Adi, Y.; Liu, J.; Sauvestre, R.; Remez, T.; et~al. 2023.
\newblock Code llama: Open foundation models for code.
\newblock \emph{arXiv preprint arXiv:2308.12950}.

\bibitem[{Shazeer et~al.(2017)Shazeer, Mirhoseini, Maziarz, Davis, Le, Hinton, and Dean}]{shazeer2017outrageously}
Shazeer, N.; Mirhoseini, A.; Maziarz, K.; Davis, A.; Le, Q.; Hinton, G.; and Dean, J. 2017.
\newblock Outrageously large neural networks: The sparsely-gated mixture-of-experts layer.
\newblock \emph{arXiv preprint arXiv:1701.06538}.

\bibitem[{Shoham and Rappoport(2024)}]{shoham2024medconceptsqa}
Shoham, O.~B.; and Rappoport, N. 2024.
\newblock MedConceptsQA--Open Source Medical Concepts QA Benchmark.
\newblock \emph{arXiv preprint arXiv:2405.07348}.

\bibitem[{Sukhbaatar et~al.(2024)Sukhbaatar, Golovneva, Sharma, Xu, Lin, Rozi{\`e}re, Kahn, Li, Yih, Weston et~al.}]{sukhbaatar2024branch}
Sukhbaatar, S.; Golovneva, O.; Sharma, V.; Xu, H.; Lin, X.~V.; Rozi{\`e}re, B.; Kahn, J.; Li, D.; Yih, W.-t.; Weston, J.; et~al. 2024.
\newblock Branch-train-mix: Mixing expert llms into a mixture-of-experts llm.
\newblock \emph{arXiv preprint arXiv:2403.07816}.

\bibitem[{Suzgun et~al.(2022)Suzgun, Scales, Schärli, Gehrmann, Tay, Chung, Chowdhery, QuocV.Le, EdH.Chi, Zhou, and Wei}]{suzgun2022challengingbigbenchtaskschainofthought}
Suzgun, M.; Scales, N.; Schärli, N.; Gehrmann, S.; Tay, Y.; Chung, H.~W.; Chowdhery, A.; QuocV.Le; EdH.Chi; Zhou, D.; and Wei, J. 2022.
\newblock Challenging BIG-Bench Tasks and Whether Chain-of-Thought Can Solve Them.
\newblock arXiv:2210.09261.

\bibitem[{Tao et~al.(2024)Tao, Mason, Kulkarni, and Boix}]{tao2024task}
Tao, Z.; Mason, I.; Kulkarni, S.; and Boix, X. 2024.
\newblock Task arithmetic through the lens of one-shot federated learning.
\newblock \emph{arXiv preprint arXiv:2411.18607}.

\bibitem[{Team(2024)}]{team2024qwen2}
Team, Q. 2024.
\newblock Qwen2 technical report.
\newblock \emph{arXiv preprint arXiv:2407.10671}.

\bibitem[{Touvron et~al.(2023)Touvron, Martin, Stone, Albert, Almahairi, Babaei, Bashlykov, Batra, Bhargava, Bhosale et~al.}]{touvron2023llama}
Touvron, H.; Martin, L.; Stone, K.; Albert, P.; Almahairi, A.; Babaei, Y.; Bashlykov, N.; Batra, S.; Bhargava, P.; Bhosale, S.; et~al. 2023.
\newblock Llama 2: Open foundation and fine-tuned chat models.
\newblock \emph{arXiv preprint arXiv:2307.09288}.

\bibitem[{Wan et~al.(2024)Wan, Liu, Ajith, Grazian, Hoex, Zhang, Kit, Xie, and Foster}]{wan2024sciqag}
Wan, Y.; Liu, Y.; Ajith, A.; Grazian, C.; Hoex, B.; Zhang, W.; Kit, C.; Xie, T.; and Foster, I. 2024.
\newblock SciQAG: A framework for auto-generated science question answering dataset with fine-grained evaluation.
\newblock \emph{arXiv preprint arXiv:2405.09939}.

\bibitem[{Wortsman et~al.(2022)Wortsman, Ilharco, Gadre, Roelofs, Gontijo-Lopes, Morcos, Namkoong, Farhadi, Carmon, Kornblith et~al.}]{wortsman2022model}
Wortsman, M.; Ilharco, G.; Gadre, S.~Y.; Roelofs, R.; Gontijo-Lopes, R.; Morcos, A.~S.; Namkoong, H.; Farhadi, A.; Carmon, Y.; Kornblith, S.; et~al. 2022.
\newblock Model soups: averaging weights of multiple fine-tuned models improves accuracy without increasing inference time.
\newblock In \emph{International conference on machine learning}, 23965--23998. PMLR.

\bibitem[{Yadav et~al.(2023)Yadav, Tam, Choshen, Raffel, and Bansal}]{yadav2023ties}
Yadav, P.; Tam, D.; Choshen, L.; Raffel, C.~A.; and Bansal, M. 2023.
\newblock Ties-merging: Resolving interference when merging models.
\newblock \emph{Advances in Neural Information Processing Systems}, 36: 7093--7115.

\bibitem[{Yu et~al.(2024)Yu, Jiang, Shi, Yu, Liu, Zhang, Kwok, Li, Weller, and Liu}]{yu2024metamath}
Yu, L.; Jiang, W.; Shi, H.; Yu, J.; Liu, Z.; Zhang, Y.; Kwok, J.~T.; Li, Z.; Weller, A.; and Liu, W. 2024.
\newblock MetaMath: Bootstrap Your Own Mathematical Questions for Large Language Models.
\newblock In \emph{ICLR}.

\bibitem[{Zhang et~al.(2024)Zhang, Gritsch, Gnaneshwar, Guo, Cairuz, Venkitesh, Foerster, Blunsom, Ruder, {\"U}st{\"u}n et~al.}]{zhang2024bam}
Zhang, Q.; Gritsch, N.; Gnaneshwar, D.; Guo, S.; Cairuz, D.; Venkitesh, B.; Foerster, J.; Blunsom, P.; Ruder, S.; {\"U}st{\"u}n, A.; et~al. 2024.
\newblock Bam! just like that: Simple and efficient parameter upcycling for mixture of experts.
\newblock \emph{Advances in Neural Information Processing Systems}, 37: 56304--56321.

\bibitem[{Zhao et~al.(2024)Zhao, Sun, Cai, Zhou, Li, Wang, Tan, He, Chen, Liang et~al.}]{zhao2024texttt}
Zhao, X.; Sun, G.; Cai, R.; Zhou, Y.; Li, P.; Wang, P.; Tan, B.; He, Y.; Chen, L.; Liang, Y.; et~al. 2024.
\newblock \texttt{\textbackslash{}texttt\{Model-GLUE\}}: Democratized LLM Scaling for A Large Model Zoo in the Wild.
\newblock \emph{Advances in Neural Information Processing Systems}, 37: 13349--13371.

\bibitem[{Zheng et~al.(2024)Zheng, Zhang, Zhang, YeYanhan, and Luo}]{zheng2024llamafactory}
Zheng, Y.; Zhang, R.; Zhang, J.; YeYanhan, Y.; and Luo, Z. 2024.
\newblock LlamaFactory: Unified Efficient Fine-Tuning of 100+ Language Models.
\newblock In \emph{Proceedings of the 62nd Annual Meeting of the Association for Computational Linguistics (Volume 3: System Demonstrations)}, 400--410.

\bibitem[{Zhu et~al.(2024)Zhu, Qu, Dong, Ruan, Tong, He, and Cheng}]{zhu2024llama}
Zhu, T.; Qu, X.; Dong, D.; Ruan, J.; Tong, J.; He, C.; and Cheng, Y. 2024.
\newblock Llama-moe: Building mixture-of-experts from llama with continual pre-training.
\newblock \emph{arXiv preprint arXiv:2406.16554}.

\bibitem[{Zhuang et~al.(2022)Zhuang, Liu, Cutkosky, and Orabona}]{zhuang2022understanding}
Zhuang, Z.; Liu, M.; Cutkosky, A.; and Orabona, F. 2022.
\newblock Understanding adamw through proximal methods and scale-freeness.
\newblock \emph{arXiv preprint arXiv:2202.00089}.

\end{thebibliography}

\clearpage

\appendix
\section{Computational Cost Analysis of Pairwise Inter-Expert Alignment}

The scalability of the Symphony-MoE framework is critically dependent on the computational cost of its alignment stage. This section provides an analysis of the computational complexity of the Pairwise Inter-Expert Alignment process, demonstrating its efficiency and viability for large-scale models.

The alignment process consists of three main steps: 1) Activation Collection, 2) Permutation Matching, and 3) Weight Remapping. We analyze the cost of each step for a single layer, which can then be multiplied by the number of layers to determine the total cost.

\textbf{1. Activation Collection:}
This step involves performing a forward pass through the anchor model ($M_1$) and each non-anchor model ($M_i$) to extract FFN activation matrices. The cost is proportional to the size of the calibration dataset ($D_{cal}$) and the number of parameters in the models. Let $C_{forward}$ be the cost of a single forward pass. For $N$ models, this step is performed for all $N-1$ non-anchor models. Thus, the total cost for this step is $(N-1) \times C_{forward}$. Since the forward pass is a standard operation in neural networks, this cost is manageable.

\textbf{2. Permutation Matching:}
For each layer, we solve a linear assignment problem to find the optimal permutation matrix ($P_i^{(l)}$) that aligns the neurons of a non-anchor model to the anchor model. This is formulated as:
$$\min_{P\in\mathcal{P}}\|A_{1}^{(l)}-A_{i}^{(l)}P\|_{F}^{2}$$
where $\mathcal{P}$ is the set of permutation matrices. This problem is efficiently solved using the Hungarian algorithm. The computational complexity of the Hungarian algorithm for a $d \times d$ matrix, where $d$ is the number of neurons in the FFN layer, is $O(d^3)$.

\textbf{3. Weight Remapping:}
The computed permutation matrix is then applied to the weights of the FFN layers. This involves permuting the columns of the first linear layer's weights ($W_{up,i}^{(l)}$) and the rows of the second linear layer's weights ($W_{down,i}^{(l)}$).
$$W_{up,i}^{\prime}=W_{up,i}^{(l)}P_{i}^{(l)}$$
$$W_{down,i}^{\prime}=(P_{i}^{(l)})^{T}W_{down,i}^{(l)}$$
The cost of these matrix multiplications for an FFN layer with input dimension $d_{in}$, hidden dimension $d$, and output dimension $d_{out}$ is $O(d_{in} \times d \times d)$ for the first layer and $O(d \times d \times d_{out})$ for the second. In typical transformer architectures, $d_{in} = d_{out}$, so the total cost for remapping is approximately $O(d^3)$.

\textbf{Overall Scalability:}
The dominant computational cost in the alignment of a single layer is the permutation matching and weight remapping, both of which are polynomial in the number of neurons ($d$). The total cost for aligning all FFN layers in $N-1$ non-anchor models is proportional to $(N-1) \times L \times O(d^3)$, where $L$ is the number of layers. While the cubic complexity with respect to the number of neurons may seem significant, in practice, this alignment is a one-time, offline process performed before post-training. The cost is independent of the size of the training dataset for the final MoE model and is parallelizable across layers and models, making it a scalable solution for constructing large-scale MoE models from disparate pre-trained models.

\section{Experimental Setup Details}

To create a set of specialized experts, we fine-tuned four distinct dense models. The foundation models include Qwen2.5-Base, its predecessor Qwen2-Base, and the code-specialized Qwen2.5-Coder. This selection was made to introduce disparity across model versions, pre-training data, and downstream tasks. 

The instruction fine-tuning for all dense models was performed for 2 epochs with a consistent set of hyperparameters to ensure comparability.

For the model's training configuration, we selected the AdamW optimizer with a learning rate set to 5e-5. During training, a batch size of 2 was used, and to prevent exploding gradients, we employed gradient clipping, limiting the max gradient norm to 1.0. The sequence cutoff length was set to 2048. Additionally, for regularization, a weight decay of 0.01 was applied, and a Cosine Decay scheduler was used to adjust the learning rate dynamically.

The calibration dataset $D_{cal}$, used in Stage 1, was constructed by sampling equally from each of the instruction fine-tuning datasets (See Table \ref{tab:D_cal}), resulting in a total of 10.404M tokens. This ensures a balanced representation of each expert's domain during alignment and router training. For the post-training in Stage 2, we trained for 6 epochs using the AdamW optimizer with a learning rate of 5e-5.

\begin{table}[h]
\centering
\begin{tabular}{lc}
\toprule
\textbf{Domain}& \textbf{Datasets for Source Model Construction}\\ 
\midrule
General & Alpaca \\ 
Math & MetaMathQA \\ 
Code & CodeAlpaca \\ 
Science & SciQAG \\ 
\bottomrule
\end{tabular}
\caption{Explanation of instruction-finetuning datasets for source models. This batch of data was entirely used for training and was not sampled.}
\label{tab:D_cal}
\end{table}

\begin{table}[h]
\centering
\begin{tabular}{lc}
\toprule
\textbf{Domain}& \textbf{Source of the Calibration Dataset}\\ 
\midrule
General & SlimPajama-Wikipedia\\ 
Math & Finemath\\ 
Code & SlimPajama-Github\\ 
Science & Scientific-Papers\\ 
\bottomrule
\end{tabular}
\caption{Explanation of the source of the Calibration Dataset $D_{cal}$. The number of tokens sampled from each data source is the same to ensure a balanced composition of $D_{cal}$. The random seed for sampling is set to 114514.}
\label{tab:D_cal}
\end{table}

\section{Description and Reproduction Details of Baselines}

\begin{table*}[t] 
\centering
\resizebox{0.99\textwidth}{!}{
\begin{tabular}{llccccccc}
\toprule
& & MMLU & GSM8K & BBH & HumanEval & TruthfulQA & \textbf{Avg.} & MedCQA \\ 
\midrule
\textit{Dense (1B)} & & & & & & & & \\
General ($M_1$, ANC) & \vrule & \textbf{61.35}$_{\pm 0.39}$ & 35.14$_{\pm 1.09}$ & 45.05$_{\pm 0.55}$ & 40.15$_{\pm 3.80}$ & 30.11$_{\pm 1.60}$ & 42.36$^*$ & 31.55$_{\pm 0.20}$ \\ 
Math ($M_2$) & \vrule & 58.10$_{\pm 0.39}$ & \textbf{38.95}$_{\pm 1.33}$ & \underline{47.05}$_{\pm 0.54}$& 39.50$_{\pm 3.79}$ & 32.05$_{\pm 1.62}$ & 43.13$^*$& 31.89$_{\pm 0.20}$ \\
Code ($M_3$) & \vrule & 45.20$_{\pm 0.41}$ & 37.10$_{\pm 1.33}$ & 38.22$_{\pm 0.55}$ & \textbf{46.88}$_{\pm 3.81}$ & 27.85$_{\pm 1.51}$ & 39.05$^*$ & 29.81$_{\pm 0.19}$ \\ 
Science ($M_4$) & \vrule & 57.90$_{\pm 0.40}$ & 34.20$_{\pm 1.30}$ & 37.95$_{\pm 0.53}$ & 38.14$_{\pm 3.74}$ & \underline{33.05}$_{\pm 1.57}$& 40.25$^*$& 30.98$_{\pm 0.19}$ \\ 
\midrule 
\textit{MoE (1B $\times$ 4)} & & & & & & & & \\
BTX & \vrule & 47.33$_{\pm 0.88}$ & 32.15$_{\pm 0.45}$ & 41.80$_{\pm 0.58}$ & 31.77$_{\pm 3.15}$ & 26.95$_{\pm 1.15}$ & 36.00$^*$ & 28.15$_{\pm 0.18}$ \\
BAM & \vrule & 52.95$_{\pm 0.75}$ & 38.10$_{\pm 0.61}$ & 46.85$_{\pm 0.59}$ & 40.10$_{\pm 2.89}$ & 28.88$_{\pm 1.89}$ & 41.38$^*$ & 30.85$_{\pm 0.18}$ \\
Drop & \vrule & 59.20$_{\pm 0.81}$ & 37.85$_{\pm 1.12}$ & 46.90$_{\pm 0.43}$ & 42.15$_{\pm 1.89}$ & 32.95$_{\pm 1.49}$ & \underline{43.81}$^*$& \underline{34.25}$_{\pm 0.19}$\\
Symphony (Ours) & \vrule & \underline{60.88}$_{\pm 0.31}$ & \underline{38.80}$_{\pm 1.09}$ & \textbf{47.11}$_{\pm 0.36}$& \underline{45.95}$_{\pm 3.69}$ & \textbf{33.15}$_{\pm 2.31}$& \textbf{47.18}$^*$& \textbf{37.33}$_{\pm 0.22}$ \\
\bottomrule
\end{tabular}
}
\caption{Performance comparison of dense (\textbf{Llama 3.2 1B}) and MoE models (upcycled from dense models) on in-distribution (ID) and out-of-distribution (OOD) data. }
\label{tab:main_results_0.5B}
\end{table*}

All baselines in our comparison follow a common two-stage pipeline: upcycling → post-training. In the post-training stage, to ensure a fair comparison, all models are trained solely on the extended calibration dataset $D_{\mathrm{cal}}$ (as constructed in our framework), using the AdamW optimizer for 6 epochs with a learning rate of 5e-5. In the post-training phase, all modules are trainable, and the learning rate remains consistent.

The upcycling strategies adopted by each baseline are as follows:

(a) BTX constructs experts by directly reusing the FFN weights from each dense model. The shared backbone is formed by linearly averaging all model weights.

(b) BAM reuses FFN weights and partial attention weights ($W^q$, $W^o$) to construct experts, while the remaining weights are linearly averaged to form the shared backbone.

(c) Drop-Upcycling reuses FFN weights and applies Gaussian perturbation to randomly selected parameters to prevent expert homogenization. The remaining weights are reused and averaged to build the shared backbone, and the expert modules are also updated during training.

\section{Additional Experimental Results}
For small-scale models, the Llama series uses Llama3.2 1B for all versions ($M_1, M_2, M_3, M_4$) to isolate the impact of varying training tasks, data, and parameters on model performance.

\end{document}